\documentclass{bmcart}

\usepackage{amsthm,amsmath}
\usepackage[utf8]{inputenc} 

\def\includegraphics{}

\startlocaldefs
\endlocaldefs
\usepackage{stackengine}
\usepackage{multirow}
\usepackage{makecell}
\usepackage{graphicx}
\usepackage{threeparttable}
\usepackage{hyperref}

\begin{document}

\begin{frontmatter}

\begin{fmbox}
\dochead{Research}


\title{Privacy-Protecting Behaviours of Risk Detection in People with Dementia using Videos}


\author[
  addressref={aff1,aff2},                   
  email={pratik.mishra@mail.utoronto.ca}   
]{\inits{P.K.M}\fnm{Pratik K.} \snm{Mishra}}
\author[
  addressref={aff2,aff3},
  email={andrea.iaboni@uhn.ca}
    ]{\inits{A.I}\fnm{Andrea} \snm{Iaboni}}
\author[
  addressref={aff2},
  email={bing.ye@utoronto.ca}
]{\inits{B.Y}\fnm{Bing} \snm{Ye}}
\author[
  addressref={aff4},
  email={kristine.newman@ryerson.ca}
]{\inits{K.N}\fnm{Kristine} \snm{Newman}}
\author[
  addressref={aff1,aff2},
  email={alex.mihailidis@utoronto.ca}
]{\inits{A.M}\fnm{Alex} \snm{Mihailidis}}
\author[
  addressref={aff1,aff2},
  email={shehroz.khan@uhn.ca}
]{\inits{S.S.K}\fnm{Shehroz S.} \snm{Khan}}


\address[id=aff1]{
  \orgdiv{Institute of Biomedical Engineering},             
  \orgname{University of Toronto},          
  \city{Toronto},                              
  \cny{Canada}                                    
}
\address[id=aff2]{%
  \orgdiv{KITE – Toronto Rehabilitation Institute},
  \orgname{University Health Network},
  \city{Toronto},
  \cny{Canada}
}
\address[id=aff3]{%
  \orgdiv{Department of Psychiatry, Temerty Faculty of Medicine},
  \orgname{University of Toronto},
  \city{Toronto},
  \cny{Canada}
}
\address[id=aff4]{%
  \orgdiv{Daphne Cockwell School of Nursing},
  \orgname{Ryerson University},
  \city{Toronto},
  \cny{Canada}
}



\end{fmbox}


\begin{abstractbox}

\begin{abstract} 
\parttitle{Background}
People living with dementia often exhibit behavioural and psychological symptoms of dementia that can put their and others' safety at risk. Existing video surveillance systems in long-term care facilities can be used to monitor such behaviours of risk to alert the staff to prevent potential injuries or death in some cases. However, these behaviours of risk events are heterogeneous and infrequent in comparison to normal events. Moreover, analyzing raw videos can also raise privacy concerns. 

\parttitle{Purpose}
In this paper, we present two novel privacy-protecting video-based anomaly detection approaches to detect behaviours of risks in people with dementia.

\parttitle{Methods}
We either extracted body pose information as skeletons or used semantic segmentation masks to replace multiple humans in the scene with their semantic boundaries. Our work differs from most existing approaches for video anomaly detection that focus on appearance-based features, which can put the privacy of a person at risk and is also susceptible to pixel-based noise, including illumination and viewing direction. We used anonymized videos of normal activities to train customized spatio-temporal convolutional autoencoders and identify behaviours of risk as anomalies. 

\parttitle{Results}
We showed our results on a real-world study conducted in a dementia care unit with patients with dementia, containing approximately 21 hours of normal activities data for training and 9 hours of data containing normal and behaviours of risk events for testing. We compared our approaches with the original RGB videos and obtained a similar area under the receiver operating characteristic curve performance of 0.807 for the skeleton-based approach and 0.823 for the segmentation mask-based approach.

\parttitle{Conclusions}
This is one of the first studies to incorporate privacy for the detection of behaviours of risks in people with dementia. Our research opens up new avenues to reduce injuries in long-term care homes, improve the quality of life of residents, and design privacy-aware approaches for people living in the community.

\end{abstract}




\end{abstractbox}
%

\end{frontmatter}



\section*{Background}
\begin{figure}[t]
\begin{center}
    \stackunder[2pt]{\includegraphics[width=4cm]{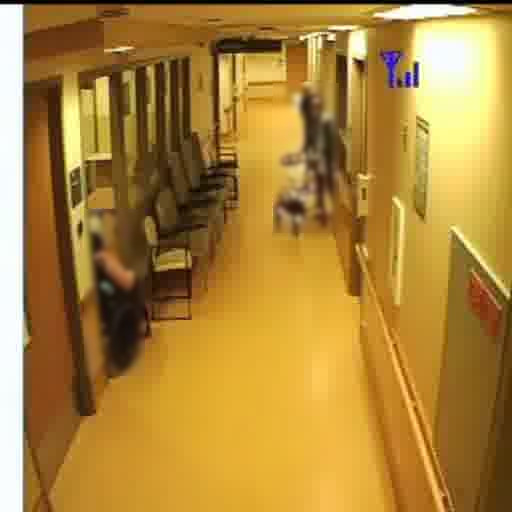}}{(a)}
    \stackunder[3pt]{\includegraphics[width=4cm]{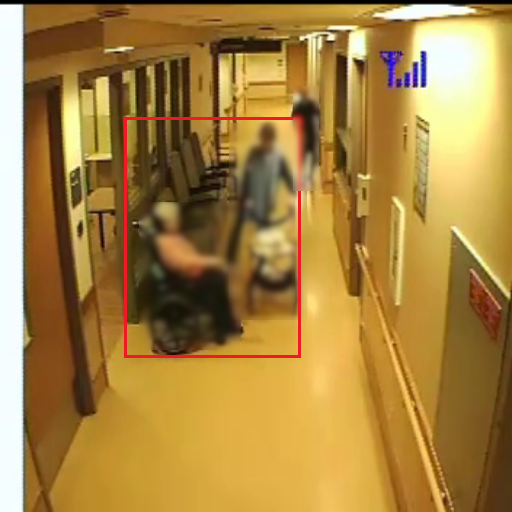}}{(b)}
    \stackunder[2pt]{\includegraphics[width=4cm]{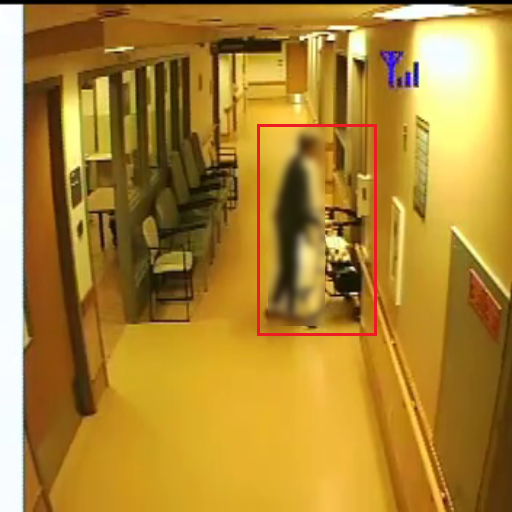}}{(c)}
\end{center}
\caption{(a) Normal event, (b) Behaviour of risk event: Patient kicking another resident on the wheelchair, (c) Behaviour of risk event: Patient banging the door. The bounding boxes are manually drawn to emphasize the behaviour of risk events. The images are blurred to protect patient privacy. }
\label{fig:events}
\end{figure}

Dementia is a syndrome that involves progressive impairment of cognitive functions such as memory and thinking and can impact the insight, impulse control and judgement of a person \cite{henderson2000definition}. It can further lead people with dementia (PwD) to exhibit behavioural and psychological symptoms of dementia, with agitation and aggression being the most common \cite{henderson2000definition}. With the progression of dementia, it becomes necessary to provide supervision and support to the PwD in their activities of daily living, which can be fulfilled by long-term care homes if home support is no longer available \cite{sloane2005evaluating}. In Canada, around 33\% of PwD younger than 80 years and 42\% of PwD 80 years or older live in long-term care homes \cite{cihidementia}. In a long-term care setting, the behaviours of risk can put PwD, other residents, and staff safety in danger. These behaviours of risk can include a range of activities related to agitation and aggression, such as hitting, kicking, punching, throwing objects, resisting care, intentional or unintentional falls, self-harm, or harm to others \cite{cohen1991instruction} (refer to Figure \ref{fig:events}). Moreover, the long-term care homes can be understaffed and lack financial resources \cite{spasova2018challenges}, which makes it difficult for the staff to monitor the PwD continuously to ensure their safety and well-being. Many care homes have video surveillance infrastructure to facilitate the digital monitoring of public spaces. However, these video camera streams are not always monitored by the staff. The feed from video cameras contain vital spatio-temporal information that can be used to develop predictive algorithms that can automatically detect the behaviours of risk events and alert clinicians or staff to enable timely intervention, thus reducing risk and health care costs and improving quality of life.

The behaviours of risk exhibited by PwD are episodic and infrequently occur in comparison to normal activities \cite{khan2019agitation}. Therefore, we propose an anomaly detection approach to identify these behaviour of risk events from the video cameras. Moreover, majority of video-based anomaly detection methods use identifiable information from individuals in the scene. This can raise privacy concerns and limit their use in residential care settings involving patients and staff \cite{rajpoot2014security, rosenfield2013patient, Senior2009}. The lack of measures to deal with the privacy of individuals can be a bottleneck in the adoption and deployment of these systems in real world \cite{climent2021protection}. One possibility to preserve privacy in videos is to extract body joints or skeleton. The existing skeleton-based approaches can utilize the compact skeleton features to identify anomalies related to the individual human postures. However, they fail to identify the anomalies related to the interaction of the individuals with each other and the objects in the environment as the skeletons only capture features related to individual human actions and motion. The behaviours of risk in PwD include different types of activities, including falls (human posture anomaly), hitting or kicking another person (human-human interaction anomaly) and destruction of property (human-object interaction anomaly). 

Considering the privacy aspect of PwD and staff and the infrequent nature of behaviours of risk events, we present novel privacy-protecting anomaly detection approaches to detect these behaviours. This paper proposes two privacy-protecting approaches for detecting behaviours of risk events in PwD as anomalies using unsupervised convolutional autoencoders using real-world video surveillance data collected from a dementia care unit. The proposed privacy-protecting approaches are based on data preprocessing steps that either extract skeletons of the individuals using human pose estimation algorithms \cite{cao2017realtime,wu2019detectron2} or use semantic segmentation \cite{chen2018encoder} to mask the appearance of the individuals. The proposed skeleton-based privacy-protecting approach involves a series of data preprocessing steps to replace the individuals in the input frames with their skeletons. This enables the convolutional autoencoders to model the body pose and actions of individuals, their interaction with each other and the objects in the environment while safeguarding their privacy.
The performance of the proposed privacy-protecting approaches is then compared with the  RGB video. \textcolor{black}{We show our results on a snapshot of approximately 30 hours of data from a larger study that collected 600 days worth of data from 17 PwD living in a care setting \cite{spasojevic2021pilot}.} Our results show that it is indeed possible to achieve an equivalent anomaly detection performance for privacy-protecting input (Area Under Curve (AUC) for Receiver Operating Characteristic (ROC) = 0.823) compared to a RGB video-based input (AUC(ROC) = 0.822) by extracting skeletons or masking the appearance of the individuals in the video frames. 

To the best of our knowledge, this is the first work that utilizes human skeletons to model human posture, human-human interaction and human-object interaction-based behaviours of risk in PwD in a privacy-protecting setting. The contributions of this paper are three-fold:
\begin{enumerate}
    \item We investigate the effectiveness of both the window and frame-level approaches corresponding to 3D and 2D convolution autoencoders, respectively, to detect the behaviour of risk events in PwD as anomalies. 
    \item We propose two privacy-protecting approaches, namely, skeleton and semantic segmentation mask-based approaches, that enable the two types of convolutional autoencoders to model the behaviours of risk in PwD related to the posture and actions of the individuals and their interaction with each other, and the objects in the environment using video surveillance data collected from a dementia care unit.
    \item We show that the proposed approaches perform equivalent to the unsupervised deep models trained on RGB videos, while protecting the appearance-based information of the people.
\end{enumerate}

The focus of this paper is to demonstrate the effectiveness of the proposed privacy-protecting approaches as an alternative and replacement to traditional RGB videos for the detection of behaviours of risk in PwD. 

\section*{Related Work}
We now present a brief overview of the existing work in the field of automatic detection of behaviours of risk in PwD using data modalities that include video. This is followed by a brief overview of the video-based anomaly detection methods that use skeletons or semantic segmentation masks to incorporate privacy in their design.

\subsection*{Behaviours of Risk Detection}
The existing work in the automatic detection of behaviours of risk, such as agitation and aggression, in PwD focuses on the use of three different sensing modalities: wearable, computer vision, and multimodal sensing. Multimodal sensing refers to a combination of wearable, and/or computer vision, and/or other ambient sensors to detect behaviours of risk in PwD. Actigraphy/accelerometer has been used previously to detect agitation and has shown correlation \cite{khan2018detecting}. Since the paper focuses on usage of videos, the further review doesn’t include accelerometer/wearable sensors and only focuses on studies that either use video alone or with other sensors. Fook et al. \cite{fook2006fusion} presented the design and implementation of a sensor fusion architecture for monitoring and handling agitation behaviour in PwD. They used ultrasound sensors, optical fibre grating pressure sensors, acoustic sensors, infrared sensors, radio-frequency identification, and video cameras in their architecture. The uncertainties of sensor measurements were modelled using Bayesian networks. Qiu et al. \cite{qiu2007multimodal} presented a multimodal information fusion approach to recognize agitation episodes in PwD. They used different modalities, namely pressure sensors, ultrasound sensors, infrared sensors, video cameras, and acoustic sensors. Low-level atomic features for agitation were extracted and a layered classification architecture was used that comprised of hierarchical hidden Markov model and support vector machine. However, the results were obtained using mock-up data created by simulation. Chikhaoui et al. \cite{chikhaoui2016ensemble} presented an ensemble learning classifier to detect agitated and aggressive behaviours using a Kinect camera and an accelerometer. Ten participants were asked to perform six agitated and aggressive behaviours. However, it was not mentioned if the participants were healthy or PwD. Fook et al. \cite{fook2007automated} presented a computer vision approach using a multi-layer architecture to identify agitation behaviour among PwD. The first layer consisted of a probabilistic classifier using Hidden Markov Models that identified decision boundaries associated with each agitation action. The output of the first layer was given as input to a discriminative classifier (called support vector machine) in the second layer to reduce inadvertent false alarms. However, the video data was of a person in bed and it was not clear if the participants were healthy or PwD. As to the best of our knowledge, this is the only work that solely used computer vision to detect agitation in PwD.

\subsection*{Skeleton-based methods}
The video-based methods operate on pixel-based appearance and motion features in videos and hence can be sensitive to noise resulting from the appearance of the individuals. Extracting information specific to the body pose of the people in the form of skeletons can help filter out the appearance-related noise for detecting abnormal events related to the posture and actions of the individuals. Human pose estimation algorithms can be used to extract body joints in the form of skeletons of the individuals in the scene \cite{cao2017realtime, fang2017rmpe}. Compared to pixel-based features, skeleton features are compact, well-structured, semantically rich, and highly descriptive about human actions and motion \cite{morais2019learning}. 
The majority of the existing skeleton-based video anomaly detection methods use the skeletons extracted for the individuals in a video frame to train a sequence \cite{morais2019learning, boekhoudt2021hr} or a graph-based \cite{markovitz2020graph, cui2021prototype} deep learning model.
Morais et al. \cite{morais2019learning} proposed a method to detect the anomalies pertaining to individual human posture and actions in surveillance videos by decomposing skeletons into two sub-components: global body movement and local body posture. The two sub-components were passed as input to a message passing gated recurrent units single-encoder-dual-decoder-based network consisting of an encoder, a reconstruction-based decoder and a prediction-based decoder. The network was trained using normal data and during testing, a frame-level anomaly score was generated by aggregating the anomaly scores of all the skeletons in a frame to identify anomalous frames.
Later, the same network was utilized for detecting crime-based anomalies in surveillance videos using pose skeletons \cite{boekhoudt2021hr}.
An unsupervised approach was proposed for detecting anomalous human actions in videos that utilized human skeleton graphs as input \cite{markovitz2020graph}. The approach utilized a spatio-temporal graph convolutional autoencoder to map the normal training samples into a latent space, which was soft assigned to clusters using a deep clustering layer.
A semi-supervised prototype generation-based graph convolutional network \cite{cui2021prototype} was proposed for video anomaly detection to reduce the computational cost associated with graph embedded networks. Pose graphs were extracted from videos and fed as input to a shift spatio-temporal graph convolutional autoencoder to learn the representation of input body joints sequences. 
Further, a semi-supervised method was proposed to jointly detect body-movement anomalies using the human posture-related features and object position-related anomalies using bounding boxes of the objects in the video frames \cite{angelini2019privacy}. However, none of the above discussed privacy-protecting video anomaly detection methods consider anomalies pertaining to human-human and human-object interactions. Our proposed approach involves passing skeletons in the form of images with the background as input to customized convolutional autoencoders to model the anomalies related to human postures as well as the interaction of people with each other and the environment.

\subsection*{Semantic segmentation-based methods}
The skeletons are a good privacy-protecting source of information about human posture. However, the quality of skeleton approximation depends upon the resolution of video frames and the degree of occlusion due to objects or people in the scene \cite{yan2020image}. Occluding the appearance of the people using semantic segmentation masks is another way to preserve the privacy of the individuals in a video frame. Similar to the skeleton-based approach, it could remove a person’s identity while maintaining the global context of the scene. Jiawei et al. \cite{yan2020image} showed that it is possible to occlude the target-related information in video frames without compromising the overall performance of human action recognition. They suggested that a model trained for human action recognition can be used to extract features for anomaly detection; however, they did not show any results on the anomaly detection task in their paper. Bidstrup et al. \cite{bidstrup2021privacy} investigated the use of semantic segmentation to maintain anonymity in video anomaly detection by transforming the individual pixels in a video frame into semantic groups. Their paper was centered around finding the best pretrained model for transforming individual pixels into semantic groups for UCHK Avenue anomaly detection dataset \cite{lu2013abnormal}. However, due to factors like view angle, color scheme, and objects in the scene, it is not clear to obtain a pretrained model that can satisfactorily transform all the pixels in a RGB frame into semantic groups for any given video dataset. Hence, in this paper, we only transform the RGB pixels for the people in the scene into semantic masks to achieve the anonymity of the individuals. When training anomaly detection methods to derive global patterns from singular pixels in RGB space, the presence of semantic boundary instead of pixels for the individuals in the scene could remove unwanted noise related to the appearance of the individuals and help the models focus on the behaviour of the individuals.

\section*{Methods}
\textcolor{black}{In this section, we describe the dataset used in this paper, the data preprocessing steps involved and the details of the convolutional autoencoders used to detect behaviours of risk in PwD.}

\subsection*{Description of Dataset}

\begin{table}[b]
    {\color{black}
    \begin{threeparttable}
    \caption{Participants’ demographic and data collection information.}
    \label{tab:info}
    \begin{tabular}{ll}
    \hline
    \#Participants                   & 17           \\ \hline
    Age (years), mean (SD)           & 78.88 (8.86) \\
    Age (years), range               & 65–93        \\
    Gender                           & Males (7)    \\
                                     & Females (10) \\
    \#Data collection days           & 600          \\
    \#Agitation days                 & 239          \\
    \#Reported agitation events      & 411          \\
    \#Fully labeled agitation events & 305          \\ \hline
    \end{tabular}
    \end{threeparttable}
    }
\end{table}

There is a scarcity of video data to study behaviours of risk in PwD in a residential care setting. The few existing approaches either use simulated environment or feasibility studies \cite{qiu2007multimodal, biswas2006agitation}. \textcolor{black}{In this paper, we utilize a novel video data on behavioural symptoms in PwD, including agitation and aggression, collected during a two-year study from 17 participants \cite{spasojevic2021pilot}. The data was collected between November 2017 and October 2019 at the Specialized Dementia Unit, Toronto Rehabilitation Institute, Canada \cite{khan2017daad}. The criterion for the recruitment of the PwD participants in the study was the exhibition of agitated behaviours in common areas of the unit. Each PwD participant was recruited in the study for a maximum of 2 months. Six hundred days' worth of video data was collected from these participants. The information related to participants' demographics and data collection are listed in Table \ref{tab:info}. A day with one or more agitation events was termed as an agitation day. The length of agitation events varied from 1 minute to 3 hours. Some agitation events were partially labeled, where the start/end time was not available. In this paper, only fully labeled agitation events (with known start and end times) are considered. Fifteen cameras were installed in public spaces (e.g., hallways, dining and recreation hall) of the dementia unit. The Lorex model MCB7183 CCD bullet camera was used, having 352x240 frame resolution, recording at 30 frames per second. Due to privacy concerns, the cameras were not installed in the bedrooms and washrooms of participating residents, and the audio was turned off. The cameras only recorded between the hours of 07:00 and 23:00.} Nurses were trained to note agitation events in their charts, which were reviewed by clinical researchers. Using this information, clinical researchers annotated the videos with agitation events manually by reviewing 15 mins before and after the reported time of the agitation events. For this paper, the behaviours of risk events from one participant and one camera was utilized. In the camera feed used for analysis, apart from the participant, other dementia residents, the staff and visitors are present. The training set comprised of approximately $21$ hours of video data, containing only normal activities, i.e., no reported agitation during that period. The test set comprised of approximately $9$ hours of video data, which consisted of the behaviour of risk events (here agitation and aggression) and 15 minutes of normal activities video data before and after the behaviour of risk events. For the test set, $22.55$ minutes out of $9$ hours of video data accounted for behaviours of risk events. Figure \ref{fig:events} shows the normal and behaviour of risk events that happened in a hallway in the unit.

\begin{figure}[t]
\begin{center}
    \stackunder[3pt]{\includegraphics[width=2.9cm, height=2.9cm]{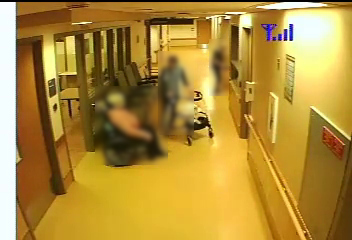}}{(a)}
    \stackunder[3pt]{\includegraphics[width=2.9cm, height=2.9cm]{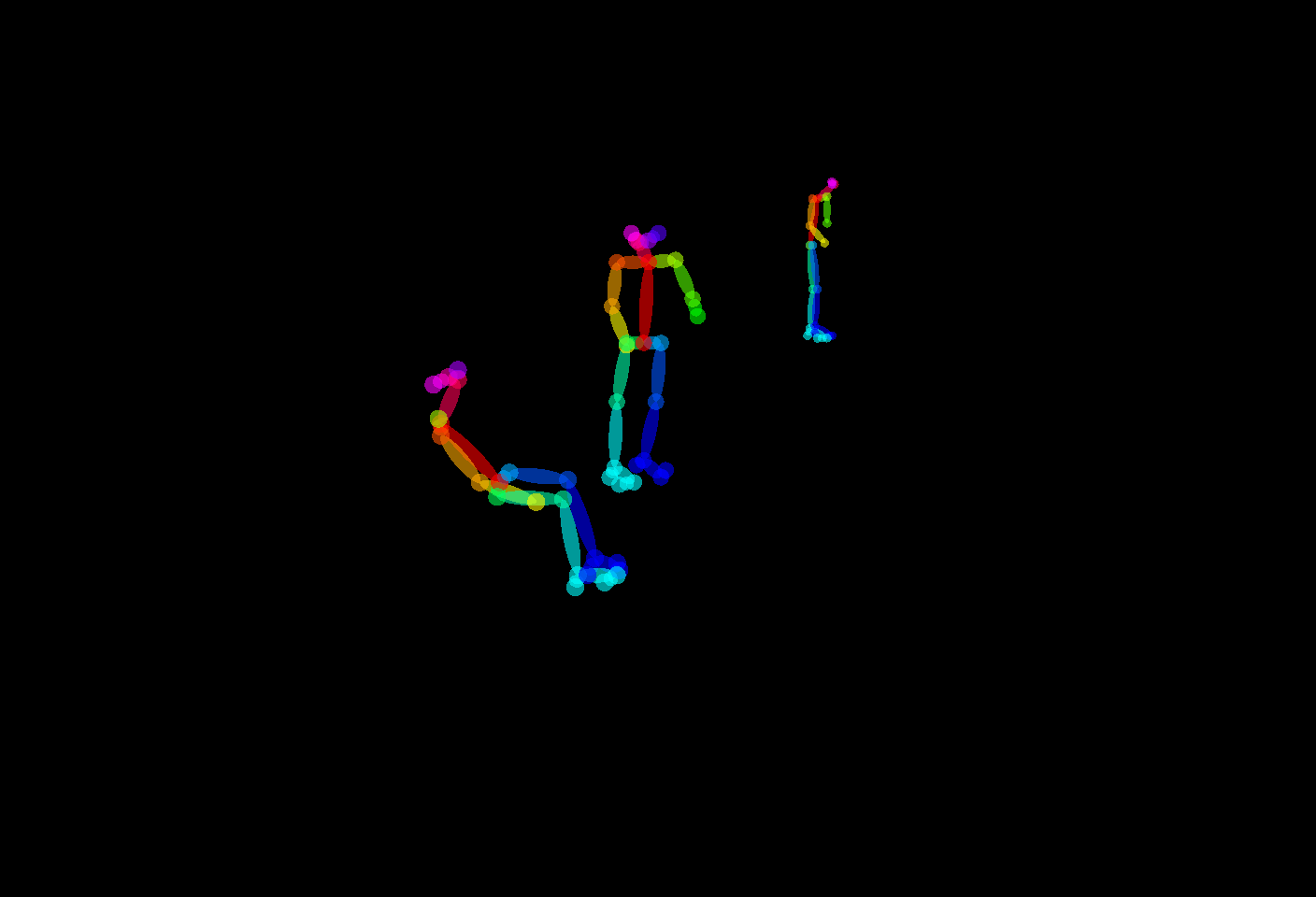}}{(b)}
    \stackunder[3pt]{\includegraphics[width=2.9cm, height=2.9cm]{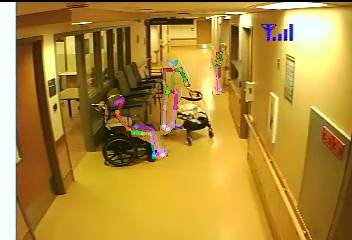}}{(c)}
    \stackunder[3pt]{\includegraphics[width=2.9cm, height=2.9cm]{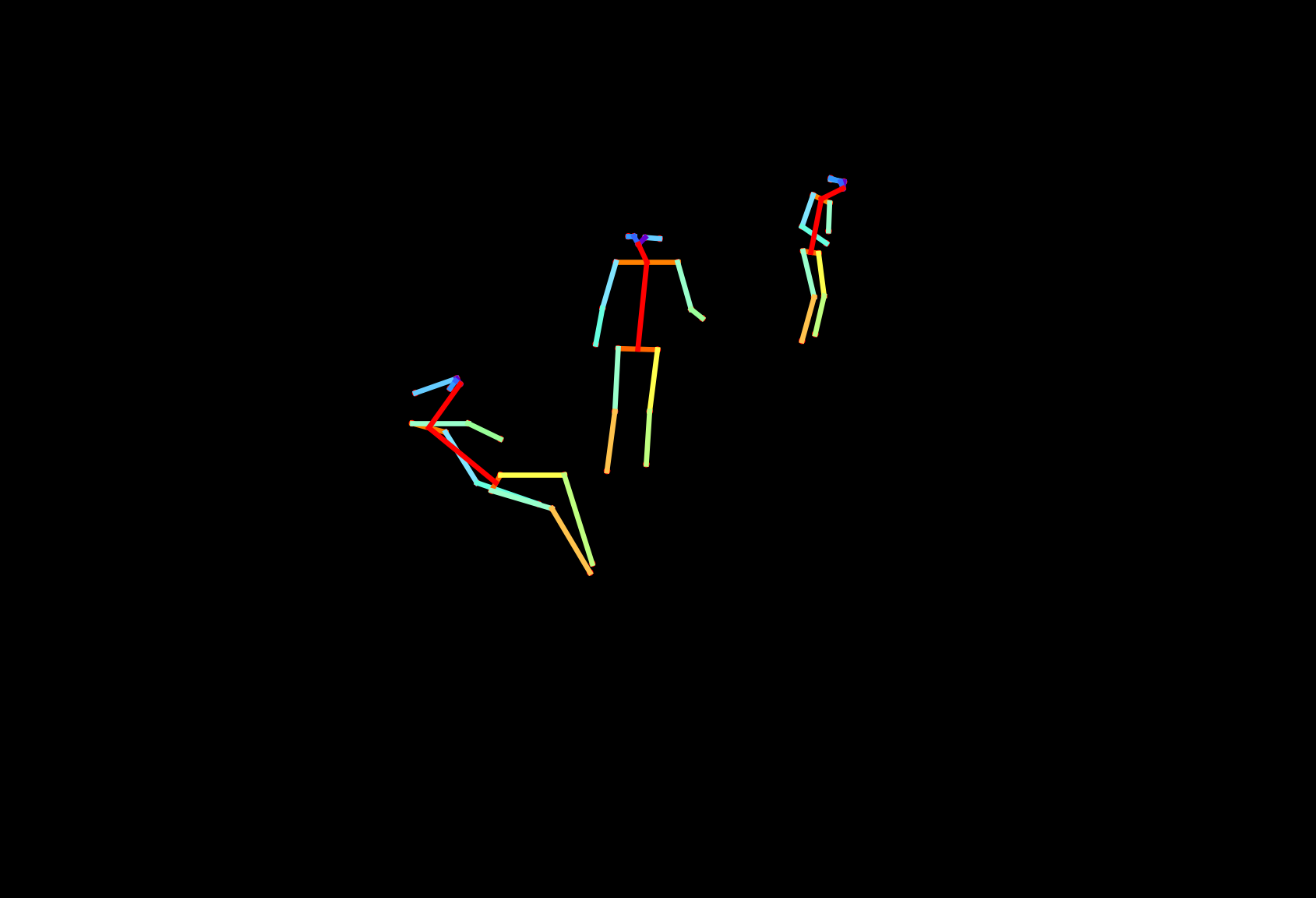}}{(d)}
    
    \stackunder[3pt]{\includegraphics[width=2.9cm, height=2.9cm]{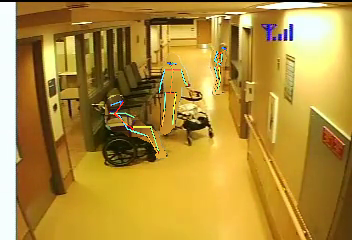}}{(e)}
    \stackunder[3pt]{\includegraphics[width=2.9cm, height=2.9cm]{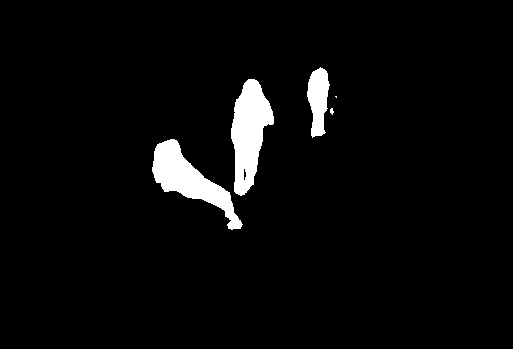}}{(f)}
    \stackunder[3pt]{\includegraphics[width=2.9cm, height=2.9cm]{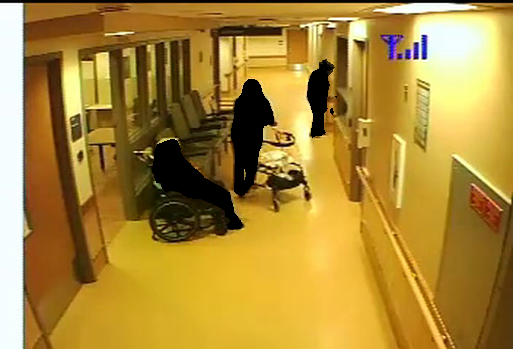}}{(g)}
\end{center}
\caption{(a) RGB video frame, (b) Openpose skeleton without background frame, (c) Openpose skeleton with background frame, (d) Detectron skeleton without background frame, (e) Detectron skeleton with background frame, (f) Segmentation mask without background frame, (g) Segmentation mask with background frame.}
\label{fig:inputs}
\end{figure}

\subsection*{Dataset Preprocessing}
The original videos had a frame rate of $30$ frames per second. \textcolor{black}{However, to ensure efficient use of computational resources, the frames were sampled at $15$ frames per second for analysis, retaining only half the frames. Oftentimes, there were presence of multiple individuals and occluding objects (i.e., carts, wheelchair, and walker) in the common areas of the unit. This made it difficult for the pose estimation algorithms to approximate the skeletons. Hence, we used two different pose estimation algorithms, namely, Openpose \cite{cao2017realtime} and Detectron2 \cite{wu2019detectron2}, for extracting skeletons for the individuals in the scene and compared their performance in identifying behaviours of risk in PwD. We created different types of privacy-protecting frames (see Figure \ref{fig:inputs}) by using various data preprocessing steps, described below.}

{\color{black}
\begin{enumerate}
    \item RGB frames: These were the RGB video frames extracted from the sampled videos, without further processing.
    \item Openpose skeleton frames without background: Openpose \cite{cao2017realtime} was used to approximate the skeletons for the participants present in each RGB frame. The appearance of the participants within the frame was then replaced with their skeletons, and the background was removed.
    \item Openpose skeleton frames with background: Openpose \cite{cao2017realtime} was used to approximate the skeletons for the participants present in each RGB frame, replacing the participants with their skeletons within the frame, while retaining the background.
    \item Detectron skeleton frames without background: Detectron2 \cite{wu2019detectron2} was used to approximate the skeletons for the participants present in each RGB frame. The appearance of the participants within the frame was then replaced with their skeletons, and the background was removed.
    \item Detectron skeleton frames with background: Detectron2 \cite{wu2019detectron2} was used to approximate the skeletons for the participants present in each RGB frame, replacing the participants with their skeletons within the frame, while retaining the background.
    \item Segmentation mask frames without background: Semantic segmentation masks \cite{chen2018encoder} depicting the participants in each RGB frame was approximated. The appearance of the participants within the frame was then replaced with their semantic masks, and the background was removed.
    \item Segmentation mask frames with background: Semantic segmentation masks \cite{chen2018encoder} was approximated for the participants present in each RGB frame, replacing the participants with their semantic masks within the frame, while retaining the background.
\end{enumerate}
}

The frames were converted to grayscale, normalized to the range $[0, 1]$ (pixel values divided by $255$) and resized to $64 \times 64$ resolution. The conversion to grayscale and resizing of the images were done to reduce the computational cost in terms of trainable parameters. The respective frames were stacked separately to form non-overlapping $5$-second windows ($75$ frames per window) to train separate convolutional autoencoders. The length of the input window was decided by the experimental analysis in our previous paper \cite{9684388}.

\subsection*{Convolutional Autoencoders}
\begin{figure*}[t]
    \footnotesize
	\begin{center}
	    \stackunder[2pt]{\includegraphics[width=0.95\textwidth]{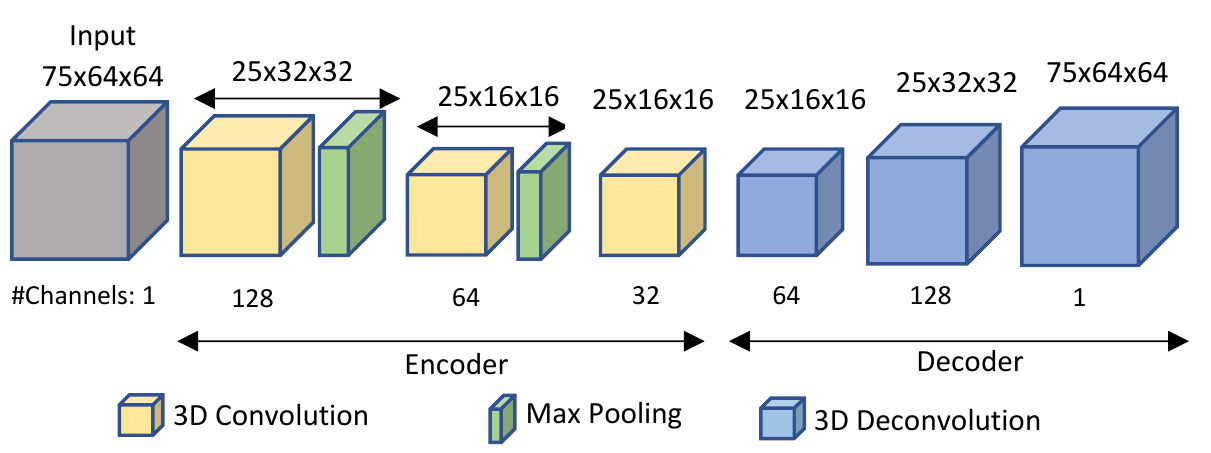}}{(a) CAE-3DConv.}
	    
	    \stackunder[2pt]{\includegraphics[width=0.95\textwidth]{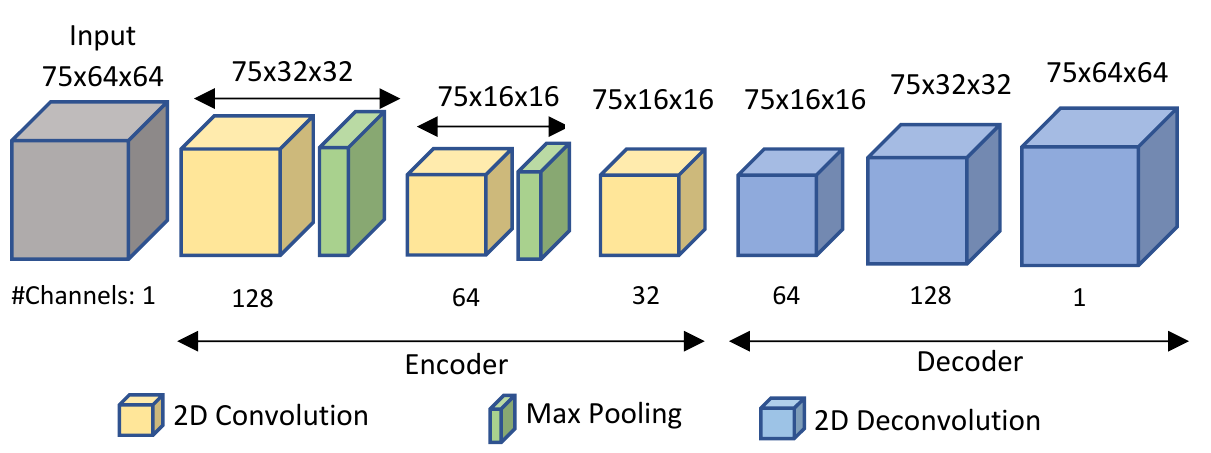}}{(b) CAE-2DConv.}
		\caption{CAE architectures to detect behaviours of risk in PwD as anomaly. (a) 3D Convolution is performed to leverage both spatial and temporal information within the input windows. (b) 2D Convolution is performed to focus solely on spatial information.}
		\label{fig:model}
	\end{center}
\end{figure*}

\textcolor{black}{Convolutional autoencoders (CAEs) learn to reconstruct the input image(s) at output by minimizing the reconstruction error during training. In general, CAEs follow an unsupervised learning approach and are trained using only normal behaviour samples. The intuition behind use of CAEs is that as they learn to reconstruct only samples representing normal behaviour during training, they are expected to give high reconstruction error for anomalous samples at test time. In existing literature, CAEs have been observed to perform well for single-scene video anomaly detection \cite{ramachandra2020survey} and extensively used for applications, such as video surveillance \cite{nawaratne2019spatiotemporal} and fall detection \cite{nogas2020deepfall}. Taking inspiration from the literature, we trained CAEs on normal videos and tested on the videos containing both normal and behaviours of risk events.} We investigated two types of approaches for training different CAEs on different privacy-protecting window inputs. The first approach was window-level, where we trained the CAE with 3D convolution (CAE-3DConv) from using previous work \cite{9684388} to leverage both spatial and temporal information in an input window. The second approach was based on frame-level, where we trained a customized CAE with 2D convolution (CAE-2DConv) to focus only on the frame-wise spatial information within an input window. Similar to CAE-3DConv, the CAE-2DConv model accepted windows as input; however, it leveraged only the spatial information within the input window by using 2D convolution to perform frame-wise reconstruction at the output. The intuition behind focusing solely on spatial information was to remove the temporal noise resulting due to movement of crowds and large objects in common areas of the dementia unit. This allowed the model to focus on the scene-based anomalies due to individual human behaviour.
The architectures for the CAE-3DConv and CAE-2DConv models are presented in Figure \ref{fig:model}. 

\subsubsection*{CAE-3DConv}
The CAE-3DConv model was adapted from the previous work by Khan et al. \cite{9684388}, and consisted of an encoder-decoder architecture, which forced the model to learn key spatio-temporal features in the input window. The encoder consisted of 3D convolution and max-pooling blocks to encode the input. The 3D convolution blocks were responsible for 3D convolution operation, followed by batch normalization and ReLU operation. A convolution kernel of size ($3\times3\times3$) with stride ($1\times1\times1$) and padding ($1\times1\times1$) was used. The first max-pooling block down sampled the spatial and temporal dimensions by a factor of 2 and 3, respectively. The second max-pooling block down sampled the spatial dimension by a factor of 2. The decoder was composed of multiple 3D deconvolution blocks, responsible for 3D transposed convolution operation followed by batch normalization. The kernel size was set to ($3\times3\times3$) with stride ($1\times1\times1$), ($1\times2\times2$), ($3\times2\times2$) and padding ($1\times1\times1$), ($1\times1\times1$), ($0\times1\times1$) for first, second, and third 3D deconvolution blocks, respectively. The parameter values were chosen to ensure that the dimensions of the output of decoder blocks match the output of the corresponding encoder blocks.

\subsubsection*{CAE-2DConv}
The CAE-2DConv model consisted of an encoder-decoder architecture, which forced the model to learn only the key spatial features in the input window. Compared to CAE-3DConv. here the encoder consisted of 2D convolution and max-pooling blocks to encode the input. The 2D convolution blocks were responsible for 2D convolution operation, followed by batch normalization and ReLU operation. A convolution kernel of size ($1\times3\times3$) with stride ($1\times1\times1$) and padding ($0\times1\times1$) was used. The spatial dimension was down sampled by a factor of 2 in the first and second max-pooling blocks. The decoder was composed of multiple 2D deconvolution blocks, responsible for 2D transposed convolution operation followed by batch normalization. The kernel size was set to ($1\times3\times3$) with stride ($1\times1\times1$), ($1\times2\times2$), ($1\times2\times2$) and padding ($0\times1\times1$), ($0\times1\times1$), ($0\times1\times1$) for first, second, and third 2D deconvolution blocks, respectively.

Both CAE-3DConv and CAE-2DConv models were trained using input windows containing only the normal activities to minimize the following reconstruction error,
\begin{equation}
    \mathcal{L}_{mse}(I, O) = \frac{1}{N_e} \sum_{l=1}^{W} \left \Vert I_l - O_l \right \Vert^2
\end{equation}
where, $I$ represents the input frames, $O$ represents the reconstructed frames, $W$ represents the number of frames in an input window (or window size), and $N_e$ is the total number of pixels in a window. In the experiments, $W=75$ and $N_e=75\times64\times64=307,200$. The intuition was that the trained model should be able to reconstruct an unseen normal window with a low reconstruction error; however, a high reconstruction error is expected for an unseen anomalous (behaviour of risk in our case) window. Hence, we used reconstruction error as an anomaly score to decide if a test window is normal or anomalous (or behaviour of risk).

\begin{table}[t]
    \begin{threeparttable}
	\caption{AUC scores for RGB and privacy-protecting inputs.}
	\label{tab:results}
	\begin{center}
	    \def\arraystretch{1.2}
        \begin{tabular}{lllll}
	    \hline
        \multirow{2}{*}{\textbf{Input Window}}                                              & \multicolumn{2}{c}{\textbf{AUC (ROC)}}                                     & \multicolumn{2}{c}{\textbf{AUC (PR)}}                                      \\ \cline{2-5} 
         & \multicolumn{1}{c}{\textbf{CAE\_3DConv}} & \multicolumn{1}{c}{\textbf{CAE\_2DConv}} & \multicolumn{1}{c}{\textbf{CAE\_3DConv}} & \multicolumn{1}{c}{\textbf{CAE\_2DConv}} \\ \hline
        \textbf{RGB}               & \multicolumn{1}{l}{0.791}  & 0.822  & \multicolumn{1}{l}{0.109}  & 0.128   \\ \hline
        \multicolumn{5}{c}{Privacy-protecting without background}  \\ \hline
        \textbf{Openpose skeleton}  & \multicolumn{1}{l}{0.763}  & 0.731  & \multicolumn{1}{l}{\textbf{0.129}}  & \textbf{0.141}   \\
        \textbf{Detectron skeleton} & \multicolumn{1}{l}{0.765}   & 0.765 & \multicolumn{1}{l}{\textbf{0.112}}       & 0.119   \\
        \textbf{Segmentation mask} & \multicolumn{1}{l}{0.640}   & 0.676   & \multicolumn{1}{l}{0.076}           & 0.117  \\ \hline
        \multicolumn{5}{c}{Privacy-protecting with background}  \\ \hline
        \textbf{Openpose skeleton}  & \multicolumn{1}{l}{\textbf{0.799}}  & 0.803  & \multicolumn{1}{l}{\textbf{0.124}}  & \textbf{0.131}   \\
        \textbf{Detectron skeleton} & \multicolumn{1}{l}{\textbf{0.807}}           & 0.812  & \multicolumn{1}{l}{\textbf{0.132}}   & \textbf{0.139}    \\
        \textbf{Segmentation mask}  & \multicolumn{1}{l}{\textbf{0.792}}           & \textbf{0.823}  & \multicolumn{1}{l}{0.100}  & 0.125  \\ \hline
        \end{tabular}
        \begin{tablenotes}
        \item[] The privacy-protecting input windows that performed better than the RGB video input are marked in bold.
        \end{tablenotes}
	\end{center}
	\end{threeparttable}
\end{table}

\section*{Results}
\begin{figure}[!t]
	\begin{center}
	    \stackunder[2pt]{\includegraphics[width=7.0cm,height=6.5cm]{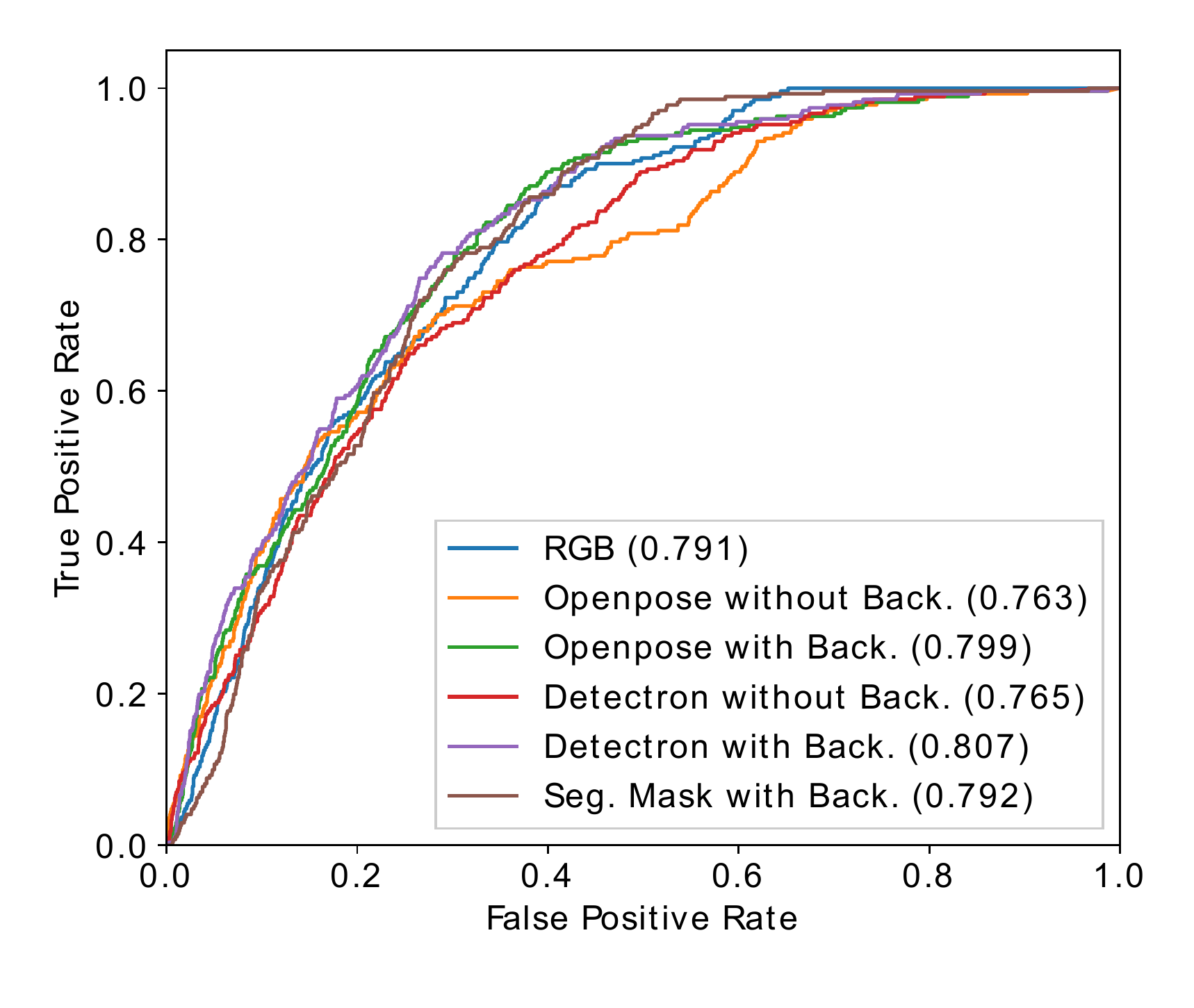}}{(a) ROC Plot.}
	    
	    \stackunder[2pt]{\includegraphics[width=7.0cm,height=6.5cm]{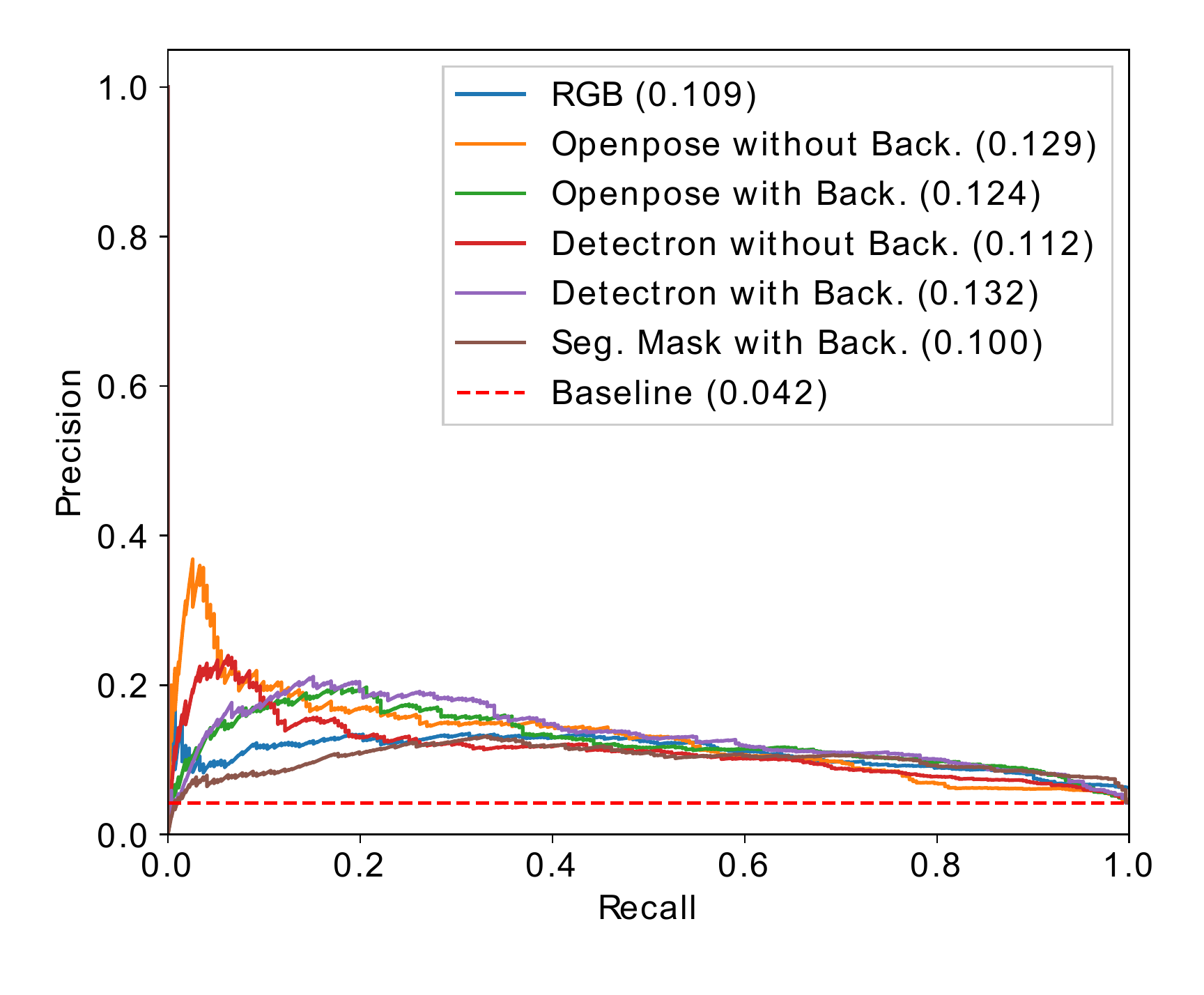}}{(b) PR Plot.}
		\caption{Comparison of curves for RGB and privacy-protecting inputs for CAE-3DConv.}
		\label{fig:auc_3D}
	\end{center}
\end{figure}

\begin{figure}[t]
	\begin{center}
	    \stackunder[2pt]{\includegraphics[width=7.0cm,height=6.5cm]{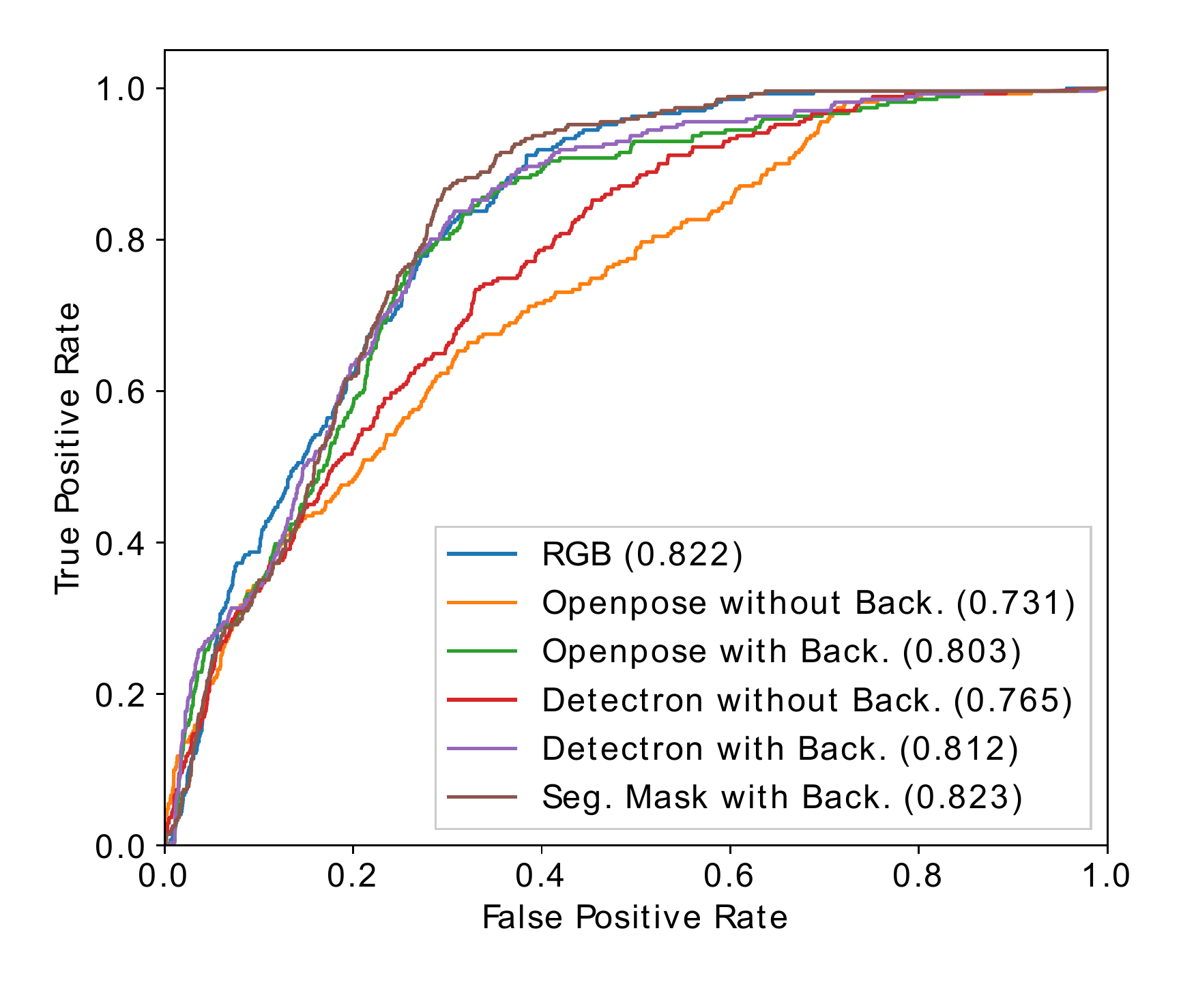}}{(a) ROC Plot.}
	    
	    \stackunder[2pt]{\includegraphics[width=7.0cm,height=6.5cm]{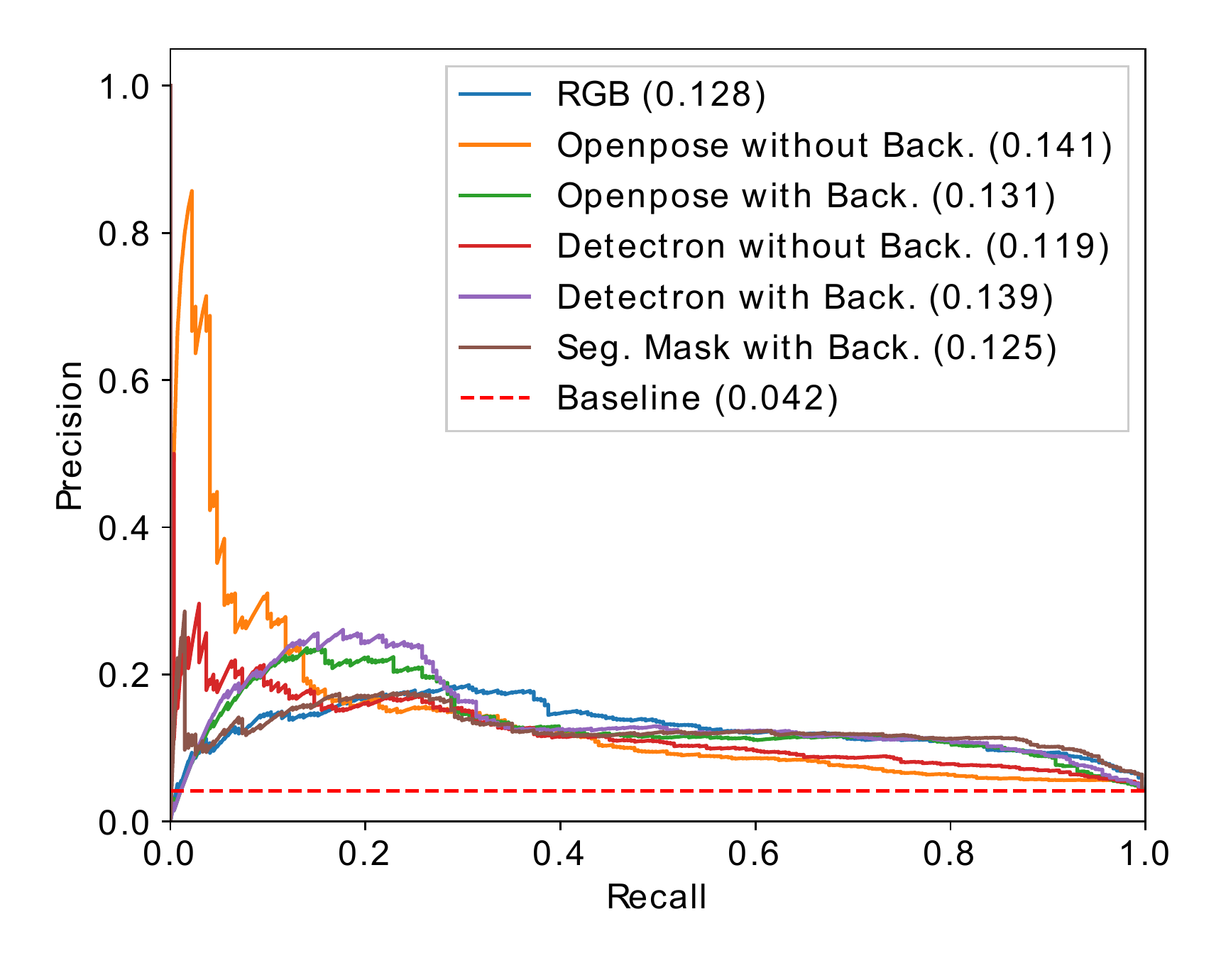}}{(b) PR Plot.}
		\caption{Comparison of curves for RGB and privacy-protecting inputs for CAE-2DConv.}
		\label{fig:auc_2D}
	\end{center}
\end{figure}

We performed experiments to investigate the effectiveness of the proposed privacy-protecting approaches in detecting behaviours of risk in PwD in comparison to RGB video inputs. We trained the CAE-3DConv and CAE-2DConv models on RGB video and different privacy-protecting inputs using the same experimental setup. Both the CAE-3DConv and CAE-2DConv models were trained for 70 epochs and used Adam optimizer with a learning rate of 0.001. The models were implemented in pytorch v1.7.1 and pytorch lightning v1.5.2 \cite{falcon2019pytorch} and run on 128 GB RAM and 32 GB NVIDIA Tesla V100 GPU CentOS 7 HPC cluster environment. The training batch size was 5, which means each batch comprised 5 windows. The per-window reconstruction error was used as an anomaly score with behaviours of risk as the class of interest. The AUC of ROC and Precision-Recall (PR) curve were used as the evaluation metrics due to the high imbalance in the test set. Table \ref{tab:results} presents the AUC(ROC) and AUC(PR) scores for the CAE-3DConv and CAE-2DConv models for RGB window and different privacy-protecting window inputs. The privacy-protecting input approaches that performed better than the RGB video input are marked in bold in the table. Figures \ref{fig:auc_3D} and \ref{fig:auc_2D} present the corresponding ROC and PR plots for RGB window and privacy-protecting window inputs for CAE-3DConv and CAE-2DConv models, respectively. In summary, the segmentation mask with background approach performed best (AUC(ROC)=0.823) among all other privacy-protecting approaches and is equivalent to the RGB-based approach (AUC(ROC)=0.822). A detailed analysis of the results is presented below. 

\begin{itemize}
    \item Table \ref{tab:results} shows that the privacy-protecting with background approaches performed consistently better than without background and are equivalent to the RGB video input. When the person appearance-related information is replaced with only the body posture information or the semantic boundary in the video frame, the privacy-protecting approaches performed equivalent to the RGB input-based approach. The underlying reason behind this observation is that even if the person appearance-based features are neglected, the key posture-based information or the shape of the target is still preserved by the proposed privacy-protecting approaches.
    \item The performance of the privacy-protecting without background approaches was lower in comparison to with background and the RGB video input. This can be attributed to the lack of information related to the objects in the environment. The behaviours of risk in PwD are a combination of different types of anomalous behaviours, including, human posture, human-human interaction and human-object interaction-based anomalies. The privacy-protecting approaches without background fail to model the human-object interaction-based anomalies, leading to poor performance. Particularly, the segmentation mask without background input contains only semantic boundaries of the individuals in the scene leading to the absence of sufficient information regarding the posture and interaction of the individuals with each other and the environment.
    \item The spatial information-based CAE-2DConv model performed slightly better than the spatio-temporal CAE-3DConv model, except for Openpose skeleton without background. The video surveillance data used in this research was taken from the common area of a dementia care unit. As such, there is frequent movement of a number of people within the video scene, leading to crowded scenes of people and objects moving at different paces. This makes it difficult for the methods to model the temporal information within the scenes, leading to lower performance when the temporal information within the window is leveraged.
    \item The baseline value for the PR curve, as can be seen in Figure \ref{fig:auc_3D} and \ref{fig:auc_2D}, is expressed as the ratio of the number of positive samples to the total number of samples. This value represents the behaviour of a random classifier. The low value of baseline is the result of the skewed data balance in the dataset due to the infrequent nature of the behaviour of risk events in comparison to normal activities. Both the CAE methods performed more than twice better than any random classifier ($0.049$) in terms of AUC(PR) score for various inputs. However, the overall low value of the AUC(PR) score shows the presence of false positives in the model predictions. This can be attributed to the presence of crowded scenes and uncommon large moving objects, leading to higher reconstruction errors in these cases.
\end{itemize}

From the above observations, it can be concluded that the privacy-protecting with background approaches that involve extracting only the skeleton information or masking the body region of the individuals in the video frames can both protect sensitive information and achieve an equivalent performance in comparison to RGB input. These results pave the way for furthering biomedical research in care and community settings to utilize videos without breaching the privacy of individuals in the form of their identifiable information. Further, the analysts can still infer the activities in the scene from the segmentation masks/skeletons.

Our approaches allow leveraging the important contextual information in the video frames while protecting the privacy of the individuals by not considering the identifiable appearance-based features. The contextual information refers to features related to the background and the interaction of the individuals with each other and the objects in the environment. The use of skeletons and segmentation masks can help to develop privacy-protecting solutions for private or community dwellings, crowded/public areas, medical settings, rehabilitation centers and long-term care homes to detect the behaviour of risk events in PwD. Cameras, such as `Sentinare 2' from Altumview \cite{altumview}, can directly extract skeletons from the humans in the scene eliminating the need to store the RGB videos in the first place. This can further ensure the protection of the privacy of the individuals.

\section*{Conclusions and Future Work} \label{conc}
Providing care for PwD in care settings is challenging due to the increasing number of patients and understaffing issues. Untoward incidents may happen in these facilities that can put the health and safety of patients, staff, and caregivers at risk. Utilizing existing video infrastructure can lead to the development of novel deep learning approaches to detect these behaviours or risk events, prevent injuries and improve patient care. However, RGB videos contain identifiable information, and their use is not straightforward in a healthcare setting. In this work, we proposed two privacy-protecting approaches for detecting the behaviours of risks in PwD, an application, where safeguarding the privacy of the individuals is a major concern. The proposed approaches are based on either extracting body postures in the form of skeletons for the people or using semantic segmentation to mask the body areas of the people in the video scenes. The privacy-protecting inputs were passed as image input to two types of convolutional autoencoders that learned the characteristics of normal video scenes and identified behaviours of risk scenes as anomalies. We investigated both window and frame-level approaches for behaviours of risk detection as anomalies using convolutional autoencoders with 3D and 2D convolutions, respectively. \textcolor{black}{We demonstrated that the privacy-protecting approaches based on skeletons (AUC(ROC)=0.812) and semantic segmentation (AUC(ROC)=0.823) with background information are able to detect behaviours of risk in PwD as anomalies with a similar performance in comparison to the RGB video input (AUC(ROC)=0.822). Hence, the skeletons and semantic masks may be viable substitutes for the appearance-based information of the people in the scene and can help preserve their privacy.}

From a clinical perspective, this work is an important step towards developing video-based privacy-protecting behaviours of risk detection system in long-term care, residential care and mental health inpatient settings. An anomaly detection framework is helpful in this regard as the behaviours of risk encompass a wide range of actions, such as falls, hitting, banging on the door or throwing furniture. In addition, it does not need the appearance characteristics of the individuals. However, the challenges in this approach are that any unusual or infrequent event, such as large moving objects or crowded scenes, could be flagged as events of interest, leading to increased false positives. A clinical monitoring system based on this technology will need to have methods in place to avoid disruptions due to these false positive alarms. Our future work includes investigating active learning approaches to reduce false positives while training the autoencoders. Further, a multi-modal approach will be investigated that uses privacy-protecting input modalities like skeletons, optical flow maps or semantic masks.


\begin{backmatter}

\section*{Abbreviations}
PwD: People with Dementia; AUC: Area Under Curve; ROC: Receiver Operating Characteristic; PR: Precision-Recall; CAE: Convolutional Autoencoder; CAE-3DConv: Convolutional Autoencoder with 3D convolution; CAE-2DConv: Convolutional Autoencoder with 2D convolution.

\section*{Declarations}
\subsection*{Ethics approval and consent to participate}
This study was approved by the research ethics board at University Health Network (REB 14-8483). Substitute decision-makers provided written consent on behalf of the PwD. The staff also provided written consent for video recording in the unit.

\subsection*{Consent for publication}
Not applicable.

\subsection*{Availability of data and materials}
Due to ethics restriction, the data may not be made available to researchers outside the institution.

\subsection*{Competing interests}
The authors declare that they have no competing interests.

\subsection*{Funding}
The project was funded through AGE-WELL NCE Inc, Alzheimer's Association, NSERC and Walter and Maria Schroeder Institute for Brain Innovation and Recovery.

\subsection*{Authors' contributions}
P.K.M presented the ideas, designed and conducted relevant experiments in the manuscript, and wrote the manuscript. A.I and S.S.K are responsible for guiding the idea and final review of the manuscript. A.I, B.Y, K.N and S.S.K collected the data used for the experiments. All authors contributed to revising the manuscript. All authors read and approved the manuscript.

\subsection*{Acknowledgements}
The authors would like to thank Robin Shan, Program Services Manager, Specialized Dementia Unit, Toronto Rehabilitation Institute, in facilitating the study and providing with the necessary logistics support. The authors express their gratitude to all the people with dementia and their families and the staff on the unit for taking part in the study.




\bibliographystyle{bmc-mathphys} 
\bibliography{ms}      






\end{backmatter}
\end{document}